\documentclass[11pt]{article}
\usepackage{coling2020}
\usepackage{times}
\usepackage{url}
\usepackage{latexsym}
\usepackage{amsmath}
\usepackage{amssymb}
\usepackage{appendix}
\usepackage{graphicx}
\usepackage{multirow}
\usepackage{bm}
\usepackage{enumitem}
\usepackage{xcolor}
\usepackage{hyperref}

\setlist{noitemsep,topsep=0pt}

\colingfinalcopy % Uncomment this line for the final submission

\title{Understanding Unnatural Questions Improves Reasoning over Text}

\author{
  Xiao-Yu Guo \\
  \\
  {\tt } \\
  \And
  Yuan-Fang Li \\
  Faculty of Information Technology, Monash University, Melbourne, Australia \\
  {\tt \{xiaoyu.guo, yuanfang.li, gholamreza.haffari\}@monash.edu} \\
  \And
  Gholamreza Haffari \\
  \\
  {\tt } \\}

\date{}

\begin{document}
\maketitle
\begin{abstract}
  Complex question answering (CQA) over raw text is a challenging task. 
  A prominent approach to this task is based on the \emph{programmer-interpreter} framework, where the \emph{programmer} maps the question into a sequence of  reasoning actions which is then executed on the raw text by the \emph{interpreter}.
  Learning an effective CQA model requires large amounts of human-annotated data, consisting of the ground-truth sequence of  reasoning actions,  which is time-consuming and expensive to collect at scale. 
  In this paper, we address the challenge of learning a high-quality programmer (parser) by projecting natural human-generated questions into unnatural machine-generated questions which are more convenient to parse. 
  %for which the sequence of actions is known. 
  %
%  a projection approach, involving mapping . 
%  With our method, it becomes possible to generate high-quality synthetic data for training models for the complex question answering task.
  We firstly generate synthetic (question, action sequence) pairs by a data generator, and train a semantic parser that associates synthetic questions with their corresponding action sequences. 
   To capture the diversity when applied to natural questions, we learn a projection model to map natural questions into their most similar unnatural questions for which the parser can work well. %can be prased by the underlying parser. 
   %finds the most similar unnatural question to a that can find a mapping from natural-language questions into synthetic questions. 
  Without any natural training data, our projection model provides high-quality action sequences for the CQA task. 
  Experimental results show that the QA model trained exclusively with synthetic data generated by our method outperforms its state-of-the-art counterpart trained on human-labeled data.
\end{abstract}

%!TEX ROOT = ./coling2020.tex

\section{Introduction}
\label{sec:introduction}
The complex question answering (CQA) task, which requires multi-step, discrete actions to be executed over text to obtain answers, is a challenging task. 
On the recently released DROP benchmark~\cite{DROP:2019}, the state-of-the-art method Neural Module Networks (NMNs)~\cite{NMNs:2019} learns to interpret each question as a sequence of neural modules, or discrete actions, and execute them to yield the answer.
The training of CQA models such as NMNs requires a large amount of (question, action sequence) pairs, which is expensive to acquire and augment. 
Therefore, the label scarcity problem remains a challenge to the CQA problem.

Motivated by this, we propose a projection model to alleviate the label scarcity challenge by generating synthetic training data.
The projection model can automatically label large amounts of unlabelled questions with action sequences,  so that a CQA model can be trained without natural supervised data. 
Our method is inspired by the recent ``simulation-to-real'' transfer approach~\cite{SimEnv:2015,UNLP:2020}. 
As the name suggests, internal knowledge in synthetic data is firstly learned in the ``simulation'' phase, and then transferred ``to-real'' circumstances with natural-language utterance using the projection model. 
In the ``simulation'' phase, we design an $n$-gram based generator to produce synthetic training data, i.e. (question, action sequence) pairs which constitute the synthetic dataset. 
We then train a semantic parser to learn the internal knowledge in this dataset. 
In the ``to-real'' phase, we train a projection model that projects each unlabelled natural-language question to a synthetic question and obtain the corresponding action sequence by interpreting the synthetic question using the trained semantic parser. 
In this way, the internal knowledge is implicitly transferred from ``simluation'' phase ``to-real'' phase.
With action sequences obtained for the natural questions, an interpreter is employed to execute these action sequences and generate answers consequently.

Experimental results on the challenging DROP dataset demonstrate the effectiveness and practicability of our projection model. 
Based on the synthetic dataset produced by the data generator, our projection model help NMNs model achieve a 78.3 F1 score on the DROP development dataset.
The result indicates the promise of ``simulation-to-real'' as a development paradigm for NLP problems.

Our contributions are as follows:
\begin{itemize}
    \item We leverage a projection model to connect synthetic questions with natural-language questions and alleviate the label scarcity problem for the CQA task. 
    \item Our projection model helps generate answers for CQA task. With high-quality action sequences, the NMNs model achieves higher F1 and Exact Match scores.
\end{itemize}

%
% The following footnote without marker is needed for the camera-ready
% version of the paper.
% Comment out the instructions (first text) and uncomment the 8 lines
% under "final paper" for your variant of English.
% 
%\blfootnote{
    %
    % for review submission
    %
    %\hspace{-0.65cm}  % space normally used by the marker
    %Place licence statement here for the camera-ready version. See
    %Section~\ref{licence} of the instructions for preparing a
    %manuscript.
    %
    % % final paper: en-uk version 
    %
    % \hspace{-0.65cm}  % space normally used by the marker
    % This work is licensed under a Creative Commons 
    % Attribution 4.0 International Licence.
    % Licence details:
    % \url{http://creativecommons.org/licenses/by/4.0/}.
    % 
    % % final paper: en-us version 
    %
    % \hspace{-0.65cm}  % space normally used by the marker
    % This work is licensed under a Creative Commons 
    % Attribution 4.0 International License.
    % License details:
    % \url{http://creativecommons.org/licenses/by/4.0/}.
%}

%!TEX Root = ./coling2020.tex
\begin{figure}[t]
    \centering
    \includegraphics[width=1.0\linewidth]{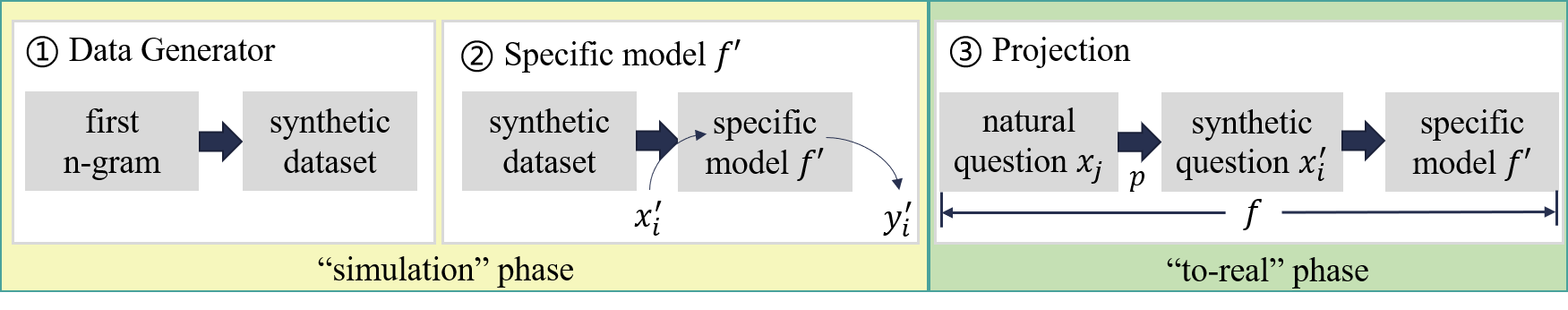}
    \caption{An overview of our proposed model. 
    \textcircled{\oldstylenums{1}} The synthetic dataset is generated by a data generator. 
    \textcircled{\oldstylenums{2}} A specific model $f'$ is trained on the synthetic dataset. 
    \textcircled{\oldstylenums{3}} A projection model is employed to project natural-language questions to synthetic questions.
    Note that \textcircled{\oldstylenums{1}} and \textcircled{\oldstylenums{2}} belong to the ``simulation'' phase, while \textcircled{\oldstylenums{3}} belongs to the ``to-real'' phase.}
    \label{fig:model}
\end{figure}

\section{Background}
\label{sec:background}
The complex question answering (CQA) task aims to generate answers for compositional questions that require a thorough understanding of questions and contents such as paragraphs.
One recent DROP dataset~\cite{DROP:2019} fits this case exactly and requires discrete reasoning over contents of paragraphs, which are extracted from Wikipedia. 
Meanwhile, all questions and corresponding answers are generated and annotated by human workers. 
The questions in DROP are complex and compositional in nature, and are thus especially challenging to existing QA models such as BiDAF~\cite{BiDAF:2016}. 

Gupta et al.~\shortcite{NMNs:2019} address this challenge with the fully differentiable Neural Module Networks (NMNs) \cite{NMNs:2015,NMNs:2016} and achieves state-of-the-art performance on DROP. 
The NMNs follows the \emph{programmer-interpreter} sturcture and  consists of two components: a parser (programmer) and an interpreter. The parser is an encoder-decoder model with attention and learns to interpret a natural-language question into an executable sequence of modules (i.e.\ actions, and thereafter we use the terms ``module'' and ``action'' interchangeably). 
The interpreter then takes the sequence along with the paragraph as inputs, and predicts the answer after executing the actions one-by-one. 

These modules, such as $find$, $filter$ and $count$, are defined to perform independent reasoning over text, numbers and dates. 
Besides, actions take arguments from questions. 
For example, the question ``How many yards was the first field goal'' is interpreted as the first action $find($field goal$)$, where ``field goal'' is the argument of the ``$find$'' action. More details can be found in Appendix \ref{app:nmns}.

For realizing the ``simulation-to-real'' projection model, we define a series of concepts as follows. 
Formally, our goal is to obtain a semantic parser, i.e.\ a \textbf{general model} $f:\mathcal{Q} \rightarrow \mathcal{T}$ that maps \emph{natural} questions $\mathcal{Q}$ to action sequences $\mathcal{T}$. 
We approximate the general model by another semantic parser, the \textbf{specific model} $f':\mathcal{Q'} \rightarrow \mathcal{T}$, where $\mathcal{Q'}$ is the set of \emph{synthetic} questions. 
Note that the model $f'$ is trained on a synthetic dataset $\mathcal{D'}=\{(x'_i, y'_i)\}$, where ($x'_i$,$y'_i$) represents a pair of (synthetic question, action sequence). 
In order to obtain the action sequence for a natural question $x_j \in \mathcal{Q}$, we propose a projection model $p:\mathcal{Q} \rightarrow \mathcal{Q'}$, with which we find a synthetic question for each natural question. 
Therefore, the action sequence for a natural question could be computed using the specific model: $f(x_j) = f'(p(x_j))$.
% In more detail, every $(x'_i, y'_i) \in \mathcal{D'}$ is generated by a first n-gram based data generator as we describe in the following section. 
The specific model $f'$ can be trained using a supervised machine learning method on synthetic data only. 
Therefore, it remains only to find the projection model $p$.

% Here, we list part of modules in Table \ref{table:modules}.
% \begin{table}[h]
%     \centering
%     \begin{tabular}{l|l|l|l}
%     \hline
%       $find$ & $find\_num$ & $year\_diff$ & $compare\_num\_lesser\_than$ \\ \hline
%       $filter$ & $find\_date$ & $year\_diff\_single\_event$ & $compare\_date\_lesser\_than$ \\\hline
%     \end{tabular}
%     \caption{Part of modules used in NMNs model.}
%     \label{table:modules}
% \end{table}

%!TEX Root = ./coling2020.tex

\section{Model}
\label{sec:model}
The overall structure of our model is shown in Figure \ref{fig:model}. 
We will firstly introduce an $n$-gram based data generator in Section~\ref{sec:generator}. 
In Section~\ref{sec:projection}, a cosine-similarity projection model and a classifier-based projection model are proposed.  
Experimental results demonstrate that the classifier-based project model achieves better question answering performance. 

\subsection{Data Generator}
\label{sec:generator}
From the existing (question, action sequence) pairs in the training set of the NMNs~\cite{NMNs:2019}, we firstly summarize a list of first $n$-grams of questions to provide sufficient coverage of the available action sequences.  
Note that the ``$n$'' in $n$-gram is a tunable parameter and there can be \textbf{multiple} action sequences for a single $n$-gram. 
Some examples are listed in Table \ref{table:n-grams}. 
\begin{table}[h]
\begin{center}
\begin{tabular}{l|l|l}
\hline \bf First $n$-grams & \bf Synthetic Questions & \bf Action Sequences \\ \hline
\multirow{3}{*}{\shortstack{how many touchdowns \\were scored}} & How many touchdowns were scored? & $count \rightarrow find$ \\ \cline{2-3}
 & \multirow{2}{*}{\shortstack{How many touchdowns were scored \\by Elam in the first quarter?}} & \multirow{2}{*}{$count \rightarrow filter \rightarrow find$} \\ 
 & & \\ \hline
\multirow{2}{*}{what happened first} & \multirow{2}{*}{\shortstack{What happened first, the crisis or the \\French Revolution?}} & \multirow{2}{*}{\shortstack{$compare\_date\_lesser\_than$ \\ $\rightarrow find\_date, find\_date$}} \\ 
 & & \\ \hline
 
% \multirow{4}{*}{how many years did it} & \multirow{2}{*}{\shortstack{How many years did it take for receipts \\to rise from 2,028 to 7,649?}}  & \multirow{2}{*}{\shortstack{$year\_difference $ \\$\rightarrow find\_date,find\_date$}} \\
%  & & \\ \cline{2-3}
%  & \multirow{2}{*}{\shortstack{How many years did it take for the Allies \\to take five towns from the Dutch?}} & \multirow{2}{*}{\shortstack{$year\_diff\_single\_event$ \\ $\rightarrow find\_date$}} \\ 
%  & & \\ \hline

% \multirow{2}{*}{which group is smaller} & \multirow{2}{*}{\shortstack{Which group is fewer: White people or \\African American?}} & \multirow{2}{*}{\shortstack{$compare\_num\_lesser\_than$ \\ $\rightarrow find\_num, find\_num$}} \\

% & & \\ 
\hline
\end{tabular}
\end{center}
\caption{\label{table:n-grams} Examples of first n-grams, synthetic questions and action sequences. }
\end{table}

Secondly, given an $n$-gram and action sequence, we generate a question with blanks. 
For example, for the $n$-gram ``what happened first'' and the action sequence ``$compare\_date\_lesser\_than \rightarrow find\_date, find\_date$'', we generate a synthetic question with two blanks to compare two event dates: ``what happened first, \underline{blank1} or \underline{blank2} ?". 
% All it remains is to fill in these blanks with entities chosen from the paragraph. 

Finally, the blanks in each generated question are replaced with names, events, noun phrases, constrained words, etc.\ from the corresponding paragraph.  
We extract these various types of entities from natural-language questions and paragraphs using spaCy\footnote{\url{https://spacy.io/}}.

% \hl{Finally, given the synthetic question, and the action sequence that we used when generating the synthetic question, we randomly sample 200 \{question, action sequences\} pairs generated by the data generator to verify the quality.
% As a result, we find 193 out of 200 pairs are consistent and correct, which proves that our data generator can effectively generate high-quality synthetic data.} \yf{not actually a part of the generation process, can be removed.}

\subsection{Projection model}
\label{sec:projection}
% A model trained exclusively on synthetic data inevitably faces difficulty arising from the large variability of natural language utterances. 
% Thus, it is necessary to project an unseen natural-language question to its most similar synthetic question. 
A straightforward way to project natural-language questions onto synthetic questions is Cosine similarity of question embeddings.
We leverage contextualized representations of question words and define the question embedding as the average of all word embeddings in the question. 
% For a question $x'_i \in \mathcal{Q}$, where $\mathcal{Q}$ is the synthetic question set defined in Section \ref{sec:background}, we define its embedding as:
% \begin{equation}
% \label{eq:emb}
%     emb(x'_i) = \frac{1}{|x'_i|} \sum_{w\in x'_i}{BERT(w)},
% \end{equation}
% where $|x'_i|$ is the number of words, $w$ denotes word in $x'_i$ and $BERT$ returns the contextualized word representations of $w$. 
Note that we employ the \textbf{bert-base-uncased} model \cite{BERT:2019} and define the projection model $p$ by:
\begin{equation}
\label{eq:cosine}
    p(x_j) = \mathop{\arg\min}_{x'_i \in \mathcal{X'}} \cos(emb(x_j), emb(x'_i)),
    % &cos(emb(x), emb(x')) = 1 - \frac{emb(x)^Temb(x')}{\Vert emb(x) \Vert \Vert emb(x') \Vert}.
\end{equation}
where $\cos$ represents the Cosine similarity between two vectors, $x_j$, $x'_i$ represents a natural and synthetic question separately. 
However, Equation \ref{eq:cosine} requires a large amount of computations, as the entire set of synthetic questions need to be compared for the projection of each natural-language question. 

To reduce the time complexity and also improve the performance of the projection model, we enumerate possible action sequences (without arguments) as class labels and treat the projection model as a classification problem. 
With contextualized representations, we define a new projection model:
\begin{equation}
    p(x_j) = classifier(emb(x_j)) = f(x_j).
\end{equation}

Note that with this classifier-based projection model, we no longer need to employ the specific model $f'$ to further find an action sequence for an input question. 
Instead, the project model $p$ interprets natural questions as action sequences directly.

%!TEX Root = ./coling2020.tex

\section{Experiments}
In this section, we will evaluate our model from two aspects. 
On the one hand, the performance of the two models proposed in Section \ref{sec:projection} will be evaluated. 
On the other hand, we employ the classifier projection model to generate action sequences and provide NMNs~\cite{NMNs:2019} with action sequences.

\subsection{Dataset and Models}
The experiments are performed on the subset of the DROP dataset used in NMNs~\cite{NMNs:2019}, containing approx.\ 19,500 (question, answer) pairs for training, 441 for validation and 1,723 for testing. 2,420 questions in the training set and all questions in the validation set have been manually annotated with ground-truth action sequences, which are used to train the NMNs model, which we denote as ``original'' below. 
%
% The baseline NMNs model is trained on the 2,420 questions with action sequences. 
We evaluate two variants of our method based on the training data used to train our classifier-based projection model: (1) ``synthetic'', where the projection model is trained on the 2,420 synthetic questions generated by the data generator in Section \ref{sec:generator}, and (2) ``natural'', where the projection model is trained on the 2,420 natural questions in the original dataset. 
Once trained, the projection model is applied to the remaining questions in the training set without action sequences to provide additional training data for NMNs. 
% \begin{itemize}
    % \item NMNs dataset: 19,000 question answer pairs for training, 440 for validation and 1700 for testing. It is a part of the DROP dataset. Since the testing set of DROP is hidden, the validation and testing set are both extracted from the DROP validation set. Note that 2,420 pairs in the training set and all pairs in the validation set are manually annotated with ground-truth action sequeneces.
%     \item Natural dataset: 2,420 (question, action sequence) pairs for training and 440 pairs for validation. All pairs are from the former NMNs dataset.
%     \item Synthetic dataset: 2,420 (question, action sequence) pairs for training and 440 pairs for validation. All pairs of the training set are generated by the data generator in Section \ref{sec:generator}.
% \end{itemize}
% Note that all the validation sets from the above three datasets are the same. Experimental results in Section \ref{sec:results} are all based on these datasets.

\subsection{Results}
\label{sec:results}
By learning models in Section \ref{sec:projection}, we employ the Cosine projection model on the synthetic dataset and find that it achieves an accuracy of 83.2\% on the validation set. 
Meanwhile, we trained two distinct classifiers using only synthetic dataset or natural dataset, with which we gain 93.2\% and 96.1\% respectively.
These results show that the synthetic dataset produced by the data generator is high-quality enough and the projection model trained on it is comparable to the one trained on natural dataset.

To further evaluate the projection model, we provide NMNs with classifier-based projection results to evaluate the performance on the downstream CQA task on the DROP dasatet~\cite{DROP:2019}. 
Concretely, we evaluate our ``simulation-to-real'' approach in two settings: ``synthetic'' and ``natural'', where two classifiers (Section \ref{sec:projection}) are trained on the synthetic dataset and natural dataset respectively. 
We compare our methods with the original NMNs model that is trained on the NMNs dataset. 
We denote this baseline method ``original''.

\begin{table}[h]
\begin{center}
\begin{tabular}{l||l|l|l}
\hline \bf Methods & \bf original (baseline) & \bf synthetic & \bf natural \\ \hline
F1 & 77.4 & 78.3 & 79.1 \\ 
EM & 74.0  & 74.9 & 75.9 \\
\hline
\end{tabular}
\end{center}
\caption{\label{table:NMNs} F1 and EM scores for NMNs trained on different datasets. }
\end{table}

In Table \ref{table:NMNs}, we report F1 and Exact Match (EM) scores for the CQA task. 
As shown in Table~\ref{table:NMNs}, compared with the original model, ``synthetic'' NMNs achieves a higher F1 and EM scores using action sequences generated by our projection model. 
Moreover, we find the quality of synthetic data can be further improved, as the projection model trained on the natural dataset achieves better performance. The reason why our projection model can help understand complex questions is mainly from two aspects. On the one hand, our projection model provide more supervised (question, action sequence) pairs with the parser. On the other hand, the action sequences produced by the projection model are more accurate than the generated sequences by parser. See details in \ref{app:example}. 

% To demonstrate the effectiveness of projection model, we select (question, action sequence) pairs from these datasets as showed in 

\section{Conclusion}
In this paper, we propose a projection model that only employs synthetic data to develop supervisions for real-world data. 
Experimental results show that our approach can be equivalent to supervised learning on natural dataset in performance. 
In addition, with projection results, we employ the NMNs model to solve complex question answering problem and generate answers for the DROP dataset. 
Higher F1, Excat Match scores demonstrate that our projection model can help improve the performance of the downstream CQA task and provide a good reference to relevant works. 

% \section*{Acknowledgements}

\bibliography{coling2020}
\bibliographystyle{coling}

\newpage
\begin{appendices}
\section{Appendix}
\subsection{NMNs model overview}
\label{app:nmns}
\begin{figure}[!htb]
    \centering
    \includegraphics[width=1.0\linewidth]{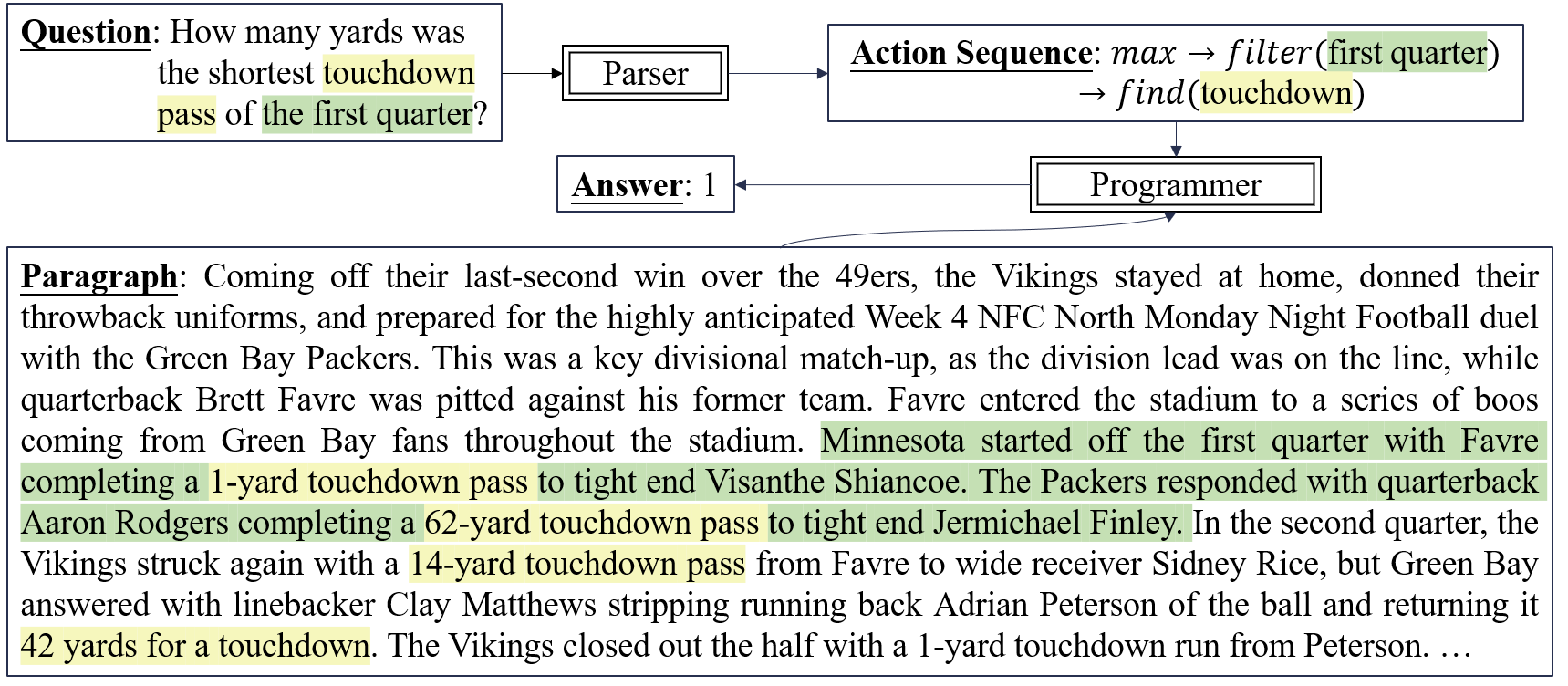}
    \caption{NMNs model architecture.}
    \label{fig:nmns}
\end{figure}

Gupta et al.~\shortcite{NMNs:2019} propose a Neural Module Networks (NMNs) model to solve the complex question answering problem. Containing a parser and an interpreter, NMNs have a more interpretable structure as shown in Figure \ref{fig:nmns}. Note that the parser and the interpreter are jointly learned with auxiliary supervision in the training period. 

As Figure \ref{fig:nmns} shows, NMNs takes as inputs the question and the paragraph. 
The encoder-decoder based parser firstly interprets the question into an executable action sequence. 
The action-based interpreter then executes this sequence against the corresponding paragraph to produce the final answer. 
By calculating attention matrix, all actions independently achieve reasoning over raw text or the outputs from other actions. 
For example, the $find($touchdown$)$ action finds all the ``touchdown pass'' in the paragraph and assigns larger attention weights to all words in yellow background. 
Then $filter($first quarter$)$ filters ``touchdown pass'' belong to the first quarter by producing an attention mask over the output of $find$ action. 
In Table \ref{table:nmns examples}, we list some examples of questions, answers and the corresponding action sequences.
\begin{table}[h]
\begin{center}
\begin{tabular}{l|l|l}
\hline \bf Questions & \bf Action Sequences & \bf Answers \\ \hline
\multirow{2}{*}{\shortstack{How many touchdowns did the Giants \\score in the fourth quarter?}} & \multirow{2}{*}{$count \rightarrow filter \rightarrow find$} & \multirow{2}{*}{2} \\ 
 & & \\ \hline
Who kicked the most field goals? & $relocate \rightarrow find$ & Rackers \\ \hline
\multirow{2}{*}{\shortstack{Who threw the longest touchdown pass \\of the first quarter?}} & \multirow{2}{*}{\shortstack{$relocate \rightarrow max$ \\ $ \rightarrow filter \rightarrow find$}} & \multirow{2}{*}{Aaron Rodgers} \\ 
 & & \\ \hline
\multirow{2}{*}{\shortstack{How many yards was the longest \\touchdown reception?}} & \multirow{2}{*}{$max \rightarrow find$} & \multirow{2}{*}{14 yards} \\ 
 & & \\ \hline
\multirow{3}{*}{\shortstack{Which happened earlier, the formation \\of the United Nations or the dissolution \\of the Soviet Union? }} & \multirow{3}{*}{\shortstack{$compare\_date\_lesser\_than$ \\$\rightarrow find\_date, find\_date$}} & \multirow{3}{*}{\shortstack{formation of the \\United Nations}} \\ 
 & & \\ & & \\ \hline
\multirow{3}{*}{\shortstack{How many years after the formation \\of the United Nations was the Universal \\Declaration of Human Rights adopted? }} & \multirow{3}{*}{\shortstack{$year\_difference \rightarrow$ \\$find\_date, find\_date$}} & \multirow{3}{*}{\shortstack{3 years}} \\ 
& & \\  & & \\ \hline
\end{tabular}
\end{center}
\caption{\label{table:nmns examples} Examples for questions, answers and action sequences.}
\end{table}

\subsection{Examples from distinct models}
\label{app:example}
Table \ref{table:examples} lists (question, action sequence) pairs generated by different projection models and compares their impact on the final answer. All questions are selected from the DROP validation dataset, and action sequences are either generated by the parser (``original'') or generated by our projection models.
As can be seen, our projection methods are able to generate action sequences that lead to correct answers. 
Apparently, the action sequences from baseline sometimes are erroneous, as shown in the first question, which does not lead to the correct answer. 
%Surprisingly, the ground-truth action sequence sometimes are erroneous, as shown in the first question, which does not lead to the correct answer. 
% For the first question, we begin to find two dates about a single event in the paragraph, and then take their difference as the answer.
% We can obtain correct action sequence thus right answer by employing projection model, but wrong answer due to incorrect action sequence when training on the original dataset. 
% However, projection model trained on synthetic dataset only achieves a limited performance. As we can see from the second example, only the projection model trained on natural dataset produce a F1 score that is greater than 0.
\begin{table}[h]
\begin{center}
\begin{tabular}{l|l|l|l}
\hline \bf Question  & \bf Method & \bf Action Sequences & \bf F1\\ \hline
\multirow{3}{*}{\shortstack{How many years did it take \\for the Allies to take five \\towns from the Dutch?}} & original & $year\_diff \rightarrow find, find$ & 0.0 \\ \cline{2-4}
 & synthetic & $year\_diff\_\bm{single\_event} \rightarrow find\_date$ & 1.0\\ \cline{2-4}
 & natural & $year\_diff\_\bm{single\_event} \rightarrow find\_date$ & 1.0\\
\hline
\multirow{6}{*}{\shortstack{Which happened first, the \\second Kandyan War, or \\Sri Lankan independence?}} & \multirow{2}{*}{original} & \multirow{2}{*}{\shortstack{$find\_span \rightarrow compare\_date\_lesser\_than$ \\ $\rightarrow find\_date, find\_date$}} & \multirow{2}{*}{0.0} \\ 
 & & & \\ \cline{2-4}
 & \multirow{2}{*}{synthetic} & \multirow{2}{*}{\shortstack{$find\_span \rightarrow compare\_date\_lesser\_than$ \\ $\rightarrow find\_date, find\_date$}} & \multirow{2}{*}{0.0}\\ 
  & & & \\ \cline{2-4}
 & \multirow{2}{*}{natural} & \multirow{2}{*}{\shortstack{$find\_span \rightarrow compare\_date\_\bm{greater}\_than$ \\ $\rightarrow find\_date, find\_date$}} & \multirow{2}{*}{0.55}\\
  & & & \\
\hline
\multirow{3}{*}{\shortstack{How many years was \\between the oil crisis and \\the energy crisis?}} & original & $year\_diff\_single\_event \rightarrow find\_date$ & 0.0 \\ \cline{2-4}
 & synthetic & $year\_diff\_single\_event \rightarrow find\_date$ & 0.0\\ \cline{2-4}
 & natural & $year\_diff \rightarrow find\_date$ & 1.0\\
\hline
\end{tabular}
\end{center}
\caption{\label{table:examples} Different (question, action sequence) pairs obtained when training on different datasets. }
\end{table}
\end{appendices}

\end{document}